\documentclass{article}
\usepackage[T1]{fontenc}
\usepackage{iclr2027_conference,times}
\iclrfinalcopy

\usepackage{amsmath,amssymb}
\usepackage{graphicx}
\usepackage[title]{appendix}
\usepackage{xcolor}
\usepackage{textcomp}
\usepackage{booktabs}
\usepackage{colortbl}
\usepackage{url}
\usepackage{etoolbox}
\usepackage[tableposition=top,skip=6pt]{caption}
\AtBeginEnvironment{thebibliography}{\raggedright}
\definecolor{linkblue}{HTML}{184581}
\usepackage[colorlinks=true,linkcolor=linkblue,citecolor=linkblue,urlcolor=linkblue]{hyperref}

\providecommand{\botrule}{\bottomrule}


\title{SUPRA: Synthesizing Amyloid PET from MRI\\and Plasma Biomarkers with an Explicit\\Regional SUVR Prior}
\author{Hanwen Li$^{1}$, Fadi Deek$^{1}$, Zhi Wei$^{1}$, Ruixiang Tang$^{2}$, Feng Liu$^{2}$, \\
\textbf{Bharat Biswal$^{1}$, Michal Schnaider Beeri$^{2}$, Auriel Willette$^{2}$, Chenxi Yuan$^{1,*}$} \\
{\normalfont $^{1}$New Jersey Institute of Technology, Newark, NJ, USA} \\
{\normalfont $^{2}$Rutgers University, New Brunswick, NJ, USA}}

\begin{document}

\maketitle
\lhead{Preprint}
{\renewcommand{\thefootnote}{}\footnotetext{$^{*}$Corresponding author: \href{mailto:chenxi.yuan@njit.edu}{\texttt{chenxi.yuan@njit.edu}}}}

\begin{abstract}
Amyloid positron emission tomography (PET) quantifies brain amyloid-$\beta$ as a
standardized uptake value ratio (SUVR), but is costly and not widely available.
Structural magnetic resonance imaging (MRI) is routine, and plasma biomarkers such as
phosphorylated tau 217 (p-tau217) are becoming practical. Synthesizing amyloid PET from
MRI alone is ill-posed, because brain anatomy reveals little about amyloid burden. We
present SUPRA (SUVR prior-guided amyloid synthesis), which divides this task between
MRI and plasma biomarkers. A regression maps the plasma panel to regional SUVR levels,
painted onto the subject's MRI segmentation as an explicit prior. A 3D network then adds
a residual correction guided by MRI. The output is in absolute SUVR, and the model also runs on p-tau217 alone. On the Alzheimer's Disease
Neuroimaging Initiative (ADNI) cohort (111 test pairs), SUPRA reaches a peak
signal-to-noise ratio (PSNR) of 23.4~dB, against at most 20.7~dB for four published
models, and rank correlations of 0.56 to 0.67 with true SUVR in the composite cortex, an
early-accumulation region and the striatum, against at most 0.24. In a second,
cognitively unimpaired cohort with other plasma platforms (143 test pairs), it reaches
22.8~dB and 0.32 to 0.61, against at most 21.4~dB and 0.20. Its area under the curve for
amyloid positivity is 0.91 and 0.82, against at most 0.63. Ablations show that plasma
supplies the amyloid information and the network the image quality. Combining MRI with a
blood test yields synthetic amyloid PET that is realistic and tracks burden across
regions.

\end{abstract}

\section{Introduction}\label{sec:intro}

Amyloid positron emission tomography (PET) is the reference method for measuring amyloid-$\beta$ (A$\beta$) in the living brain~\citep{clark2011,klunk2004}, usually quantified as the standardized uptake value ratio (SUVR). Together with cerebrospinal fluid testing, it is used to confirm
amyloid pathology before anti-amyloid treatment and to select participants for
prevention trials~\citep{vandyck2023,sperling2014}. However, amyloid PET is expensive, exposes patients to
radiation, and is not available in many hospitals~\citep{hansson2023}. Structural magnetic resonance imaging (MRI), in contrast, is part of routine clinical care. Blood tests are also becoming practical:
plasma phosphorylated tau 217 (p-tau217) now identifies amyloid-positive individuals with good
accuracy~\citep{palmqvist2020,barthelemy2024}. Yet a blood test cannot show where in the brain amyloid has
accumulated. If an amyloid PET image could be synthesized from an MRI scan and a
blood sample, this regional information would become available far more widely.

Synthesizing amyloid PET from MRI alone is an ill-posed problem. Amyloid builds up
years before brain atrophy appears, and the two are only loosely
related~\citep{jack2013}. Two brains with very similar anatomy can therefore carry very
different amounts of amyloid~\citep{jansen2015}. A network trained only on MRI tends to
predict the average uptake of its training cohort. Our evaluation on the Alzheimer's Disease Neuroimaging Initiative (ADNI) cohort shows that
published synthesis models do not yet solve this problem. We tested four of them,
three using MRI only and one also using plasma~\citep{li2024pasta,sargood2026cocolit,ou2025ilcldm,chen2025plasmacyclegan}. All four barely track
the true cortical amyloid burden, and they separate amyloid-positive from
amyloid-negative subjects poorly.

Plasma biomarkers carry the information that MRI lacks. Recent multimodal methods
have begun to combine plasma markers with MRI, typically as an additional
conditioning input to the generator~\citep{chen2025plasmacyclegan,gong2025taugennet}. These studies show that blood
information helps. However, the question of how plasma information should enter the
image, so that its contribution is explicit and its effect can be traced, has
received little attention. It also remains unclear how well such models carry over
to other cohorts and assay platforms.

Our method, SUPRA (SUVR prior-guided amyloid synthesis), is based on a simple
division of labor. The blood test indicates how
much amyloid a person carries. The MRI shows the individual brain anatomy on which
this amyloid must be placed. Following this idea, we first turn the plasma panel
into an image. A regional regression predicts the mean SUVR of each brain structure
from the plasma markers, and these values are painted onto the subject's own brain
segmentation. The result is a subject-specific SUVR prior: a coarse amyloid PET image that has the predicted level of each brain structure but no subject-specific detail. Its regression
coefficients and its regional levels can be inspected directly, so the blood-based
prediction is explicit.

A 3D network then refines this prior. It predicts only a residual correction, so it
starts from the predicted regional levels and adds the detail that the prior lacks.
MRI features guide this correction voxel by voxel at every scale. Because the
prior is in SUVR units, the output is in absolute SUVR and can be read against a positivity threshold calibrated for the target cohort (Section~\ref{sec:res_positivity}). Availability indicators let the model
run when some plasma assays are missing.

We evaluate the method on ADNI and on a second cohort, A4/LEARN, which combines the Anti-Amyloid Treatment in Asymptomatic Alzheimer's (A4) study and the Longitudinal Evaluation of Amyloid Risk and Neurodegeneration (LEARN) study. This cohort differs from ADNI in two ways: its participants were cognitively unimpaired
at enrollment, and its plasma markers were measured on different assay platforms. On
ADNI, SUPRA reaches a peak signal-to-noise ratio (PSNR) of 23.4~dB, a rank correlation of 0.67 with the true cortical burden and an area under the receiver operating characteristic curve (AUC) of 0.91 for amyloid positivity, compared with at most 20.7~dB, 0.23 and 0.63 for the published models. It also outperforms all four on A4/LEARN. In three regions from the amyloid literature (the composite cortex behind the official amyloid status, an early-accumulation region and the striatum), its rank correlations are 0.64, 0.67 and 0.56 on ADNI and 0.59, 0.61 and 0.32 on A4/LEARN, against at most 0.24 and 0.20 for the published models. Unlike a single blood value, the synthesized scan thus gives an estimate for each of these regions.

Ablation experiments support the division of labor behind the design. Without
plasma, the synthesized images look almost as good, but they lose most of their
amyloid information: the cortical correlation falls from 0.67 to 0.21 on ADNI. The
refinement network raises image quality by about 8~dB in PSNR over the prior alone,
and most of this gain remains when the network receives no MRI input. The subject's
anatomy, carried by the prior, thus shapes the image, while the MRI intensities add a smaller refinement. The method brings image quality and regional amyloid information together in one scan.

The main contributions of this work are:
\begin{itemize}
\item SUPRA, a method for amyloid PET synthesis from MRI and plasma. It turns plasma biomarkers
  into an explicit regional SUVR prior and refines this prior with an
  anatomy-guided residual network. The output is in absolute SUVR units.
\item An evaluation of image quality, amyloid positivity and regional amyloid burden, including regions affected early and late in the disease, on two cohorts that differ in population and assay platform.
\item Ablation experiments that separate the roles of the plasma panel, the MRI
  input and the refinement network.
\end{itemize}

\section{Related work}\label{sec:related}

\paragraph{PET synthesis from MRI.}
Deep generative models have been widely used to translate one medical imaging
modality into another~\citep{dayarathna2024}. Early work on MRI-to-PET synthesis used
convolutional networks and generative adversarial networks, including paired and
unpaired (CycleGAN-style) training~\citep{isola2017,zhu2017,sikka2018,pan2018}. More recent methods use diffusion
models, either in image space or in the latent space of an autoencoder~\citep{ho2020,rombach2022}.
PASTA, CoCoLIT and IL-CLDM belong to this group and are among our comparison
methods~\citep{li2024pasta,li2026pasta,sargood2026cocolit,ou2025ilcldm}. Most of these methods were developed for fluorodeoxyglucose (FDG) or amyloid PET and
were evaluated mainly with image-similarity metrics such as PSNR and the structural similarity index (SSIM)~\citep{dayarathna2024,wang2004}. For amyloid PET, these metrics say little about whether the
synthesized scan carries the correct amyloid burden. We therefore also evaluate
regional burden and amyloid positivity.

\paragraph{Non-imaging information in image synthesis.}
Several methods add non-imaging information to the synthesis network, for example
demographics, clinical scores or genetic risk~\citep{li2024pasta,ou2024prior,li2025sim2p}. This information usually
enters as a conditioning vector, through concatenation, feature modulation or
attention~\citep{perez2018,polsterl2021}. Plasma-CycleGAN follows this approach with plasma A$\beta$42/40 and
showed that blood information can improve amyloid PET synthesis~\citep{chen2025plasmacyclegan}. In
these designs, the effect of the non-imaging input is learned inside the network and
is not directly visible. A different line of work guides synthesis with an image
prior, such as an atlas or a population template, and lets the network predict
the residual~\citep{lauritzen2019}. Our method follows this second line, with an explicit regional prior image built from plasma.

\paragraph{Plasma biomarkers of amyloid pathology.}
Plasma assays of A$\beta$42/40, phosphorylated tau and markers of
neurodegeneration and astrocytic activation, such as neurofilament light chain (NfL) and glial fibrillary acidic protein (GFAP), are now
available on several platforms~\citep{janelidze2021,hansson2023}. Among them, plasma p-tau217 shows the strongest agreement with amyloid PET~\citep{janelidze2023,ashton2024,palmqvist2025}, and several p-tau217-based blood tests have recently been cleared by the US Food and Drug Administration for patients with cognitive symptoms~\citep{fda2025lumipulse,c2n2026precivityad2,roche2026elecsys}.
However, assay values differ between platforms, and cut-offs fitted on one platform
do not transfer directly to another~\citep{janelidze2021,hansson2023}. In practice, the available panel
also differs between studies and clinics. A method that uses plasma should
therefore handle missing assays and should be validated on more than one cohort and
platform. We address both points: the model accepts any subset of assays, and we
evaluate it on a second cohort with different assay platforms.

\section{Methods}\label{sec:methods}

\subsection{Overview}\label{sec:overview}

\begin{figure}[tbp]
\centering
\includegraphics[width=\textwidth]{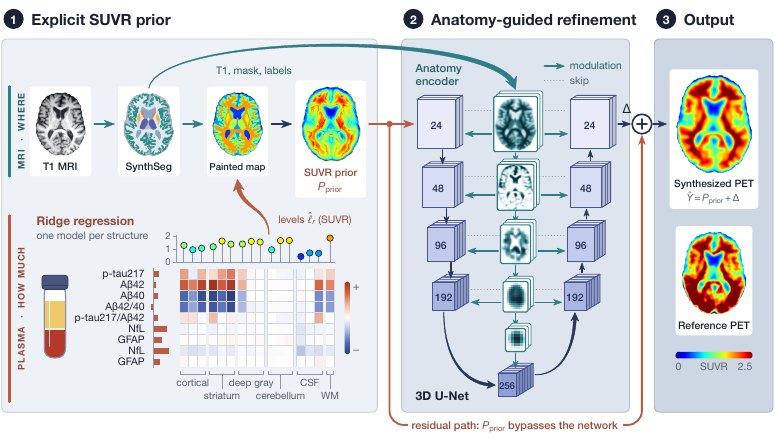}
\caption{Overview of the method. \textcircled{1}~One ridge regression per brain
structure maps the plasma panel to a predicted SUVR level (heat map: relative weights
of the nine assays for the 16 structures, orange raising and blue lowering the level;
bars: the subject's standardized assay values; dots: the predicted levels). The levels
are painted onto the subject's own SynthSeg labels, smoothed, and multiplied by a cohort
texture template, which gives the SUVR prior $P_\mathrm{prior}$. \textcircled{2}~A 3D
U-Net refines the prior (numbers: channels per level; dotted lines: skip connections).
An anatomy encoder modulates every U-Net level voxel by voxel (teal arrows).
\textcircled{3}~The output is $P_\mathrm{prior}$ plus the predicted correction $\Delta$,
in absolute SUVR. All images are from one ADNI test subject; PET-like panels share a
0--2.5 SUVR scale.}
\label{fig:method}
\end{figure}

Our goal is to synthesize an amyloid PET image, in absolute SUVR units, from a
structural MRI scan and a panel of plasma biomarkers. The method gives each input a
clear role. The plasma panel indicates how much amyloid a person carries,
and the MRI provides the anatomy on which this amyloid is placed. SUPRA has two
stages (Fig.~\ref{fig:method}). First, the plasma panel is turned into a
subject-specific SUVR prior: a coarse PET-like image whose regional levels are
predicted from the blood values (Section~\ref{sec:prior}). Second, a 3D network
refines this prior under the guidance of the subject's anatomy
(Section~\ref{sec:network}). The network predicts only a correction, and the final
image is the prior plus this correction.

\subsection{Plasma-derived regional SUVR prior}\label{sec:prior}

\paragraph{Plasma features.}
Each subject has up to nine plasma measurements (Section~\ref{sec:data}), and any of
them may be missing. Measured values are log-transformed, standardized with
training-set statistics and clipped to $[-5, 5]$. This gives a value $z_m$ for each
marker $m$. A missing value is set to $z_m = 0$, and an availability bit
$a_m \in \{0,1\}$ records whether the marker was measured. These bits let the model
tell a missing assay apart from an average value. With a one-hot code $\mathbf{t}$ of
the PET tracer, the features of a subject are
$\mathbf{x} = [\mathbf{z}, \mathbf{a}, \mathbf{t}] \in \mathbb{R}^{21}$.

\paragraph{Regional regression.}
We group the SynthSeg labels into 16 brain structures plus background.
Fig.~\ref{fig:method} shows them in six groups: cortical (cortex, hippocampus,
amygdala), striatum (caudate, putamen, nucleus accumbens), deep gray (thalamus,
pallidum, ventral diencephalon), cerebellum (cerebellar gray and white matter, shown
with the brainstem), cerebrospinal fluid (CSF; lateral ventricles, third and fourth
ventricles, extracerebral CSF) and cerebral white matter (WM). For
each structure $r$, a ridge regression predicts the log mean SUVR inside the
structure, $y_r = \log \ell_r$, from the features $\mathbf{x}$. The targets are
measured on the training PET scans. We chose ridge regression for three reasons. It
is stable with a few hundred training subjects. Its penalty can be tuned cheaply,
here by closed-form leave-one-out error over $\{0.3, 1, 3, 10, 30, 100\}$. Most
importantly, its coefficients can be read directly: each one states how much one
assay raises or lowers the predicted SUVR of one structure (Fig.~\ref{fig:method}).

\paragraph{Correcting regression shrinkage.}
Ridge regression pulls its predictions toward the mean. In our data, the predicted
cortical level varied less across subjects than the true level, and the predicted
hippocampal level varied much less. This shrinkage would carry over to the
synthesized images. We therefore rescale the centered prediction $f_r(\mathbf{x})$ of
each structure, so that the spread of its fitted values on the training set matches that of the true values:
\begin{equation}
\hat{\ell}_r = \exp\!\big(\bar{y}_r + \kappa_r f_r(\mathbf{x})\big), \qquad
\kappa_r = \operatorname{clip}_{[1,\,2]}\!\left(\frac{\operatorname{sd}(y_r)}{\operatorname{sd}(f_r)}\right),
\end{equation}
where $\bar{y}_r$ is the training mean of $y_r$. The upper bound of 2 limits how much
noise is amplified in a poorly predicted structure.

\paragraph{Painting the prior.}
The predicted levels are painted onto the subject's own label map $s(v)$, which comes
from the MRI: every voxel $v$ receives the level of its structure,
$Q(v) = \hat{\ell}_{s(v)}$. This map is smoothed with a Gaussian kernel $G$
(standard deviation of 2.5 voxels), so that structure boundaries are not hard steps.
It is then multiplied by a cohort texture template $T$: the mean training PET,
normalized to a mean of 1 inside every structure. The template adds typical contrast
within structures and leaves the regional levels nearly unchanged. Finally, voxels
outside the brain mask $M$ are set to zero:
\begin{equation}
P_\mathrm{prior} = M \odot (G * Q) \odot T.
\end{equation}
The result is a coarse PET image in SUVR units, with the subject's predicted regional
levels and anatomy but without subject-specific detail. It also makes the
plasma-based prediction explicit: before any network is applied, the prior shows
what the blood test predicts for each region of this brain.

\subsection{Anatomy-guided residual refinement}\label{sec:network}

\paragraph{Residual refinement.}
The refinement network is a 3D U-Net with five resolution levels
(Fig.~\ref{fig:method}). It receives the prior together with anatomy features of the
subject. Instead of synthesizing the PET image from scratch, it predicts a correction
$\Delta$ to the prior:
\begin{equation}
\hat{Y} = P_\mathrm{prior} + \Delta(P_\mathrm{prior}, \mathbf{F}),
\end{equation}
where $\mathbf{F}$ are the anatomy features, described below. Plasma reaches the network through the prior, which already carries the predicted regional levels. The
correction therefore only has to add what the prior lacks: detail within structures,
sharp boundaries and adjustments of the regional levels. The output layer is
initialized to zero, so training starts from the prior itself. During training, each
regional level of the prior is multiplied by its own random factor $\exp(\epsilon)$,
with $\epsilon \sim \mathcal{N}(0, 0.05^2)$, so that the network learns to correct
the prior rather than copy it. Because the prior is in SUVR units, $\hat{Y}$ is an
absolute SUVR image and needs no rescaling.

\paragraph{Anatomy guidance.}
A separate convolutional encoder extracts anatomical features from the MRI, the brain
mask and a one-hot map of the 17 label classes. It returns one feature map
$\mathbf{F}^{(k)}$ at each of the five U-Net scales, $k = 1,\dots,5$, and receives no
plasma input. The full-resolution features $\mathbf{F}^{(1)}$ are part of the U-Net
input. In addition, the features steer every residual block of the U-Net through
adaptive group normalization (AdaGN). Each group normalization (GN) in a block is
followed by a modulation:
\begin{equation}
\operatorname{AdaGN}(\mathbf{h}) =
\operatorname{GN}(\mathbf{h}) \odot \big(1 + \boldsymbol{\gamma}(\mathbf{F}^{(k)})\big)
+ \boldsymbol{\beta}(\mathbf{F}^{(k)}),
\end{equation}
where $\mathbf{h}$ are the features of a block at scale $k$. The terms $\boldsymbol{\gamma}$ and $\boldsymbol{\beta}$ are $1\times1\times1$ convolutions
of the anatomy features. They vary from voxel to voxel and act in the encoder, the
bottleneck and the decoder alike. Anatomy thus guides the correction at every scale
and at every location. All modulation layers are initialized to zero, so every block starts as an unconditioned residual block.

\subsection{Training objective}\label{sec:training}

Let $Y$ be the true PET image, and let $R_\mathrm{ctx}$ and $R_\mathrm{hip}$ be the
cortex and hippocampus within the brain mask $M$. The training loss is
\begin{equation}
\mathcal{L} = \mathcal{L}_1^{M} + 0.5\,\mathcal{L}_1 + \mathcal{L}_\mathrm{SSIM}^{M}
+ 0.25\,\mathcal{L}_\nabla^{M} + 0.5\,\mathcal{L}_\mathrm{ROI}.
\end{equation}
Here, $\mathcal{L}_1^{M}$ and $\mathcal{L}_1$ are the mean absolute errors inside the
brain and over the whole volume. The term $\mathcal{L}_\mathrm{SSIM}^{M}$ is one
minus the structural similarity inside the brain. The term $\mathcal{L}_\nabla^{M}$
is the mean absolute difference of the spatial gradients inside the brain, which
keeps structure boundaries sharp. The region-of-interest (ROI) term $\mathcal{L}_\mathrm{ROI}$ is the absolute
error of the regional mean SUVR, averaged over the whole brain, $R_\mathrm{ctx}$ and
$R_\mathrm{hip}$. It targets the regional amyloid burden directly, which the
voxel-wise terms do not. Data augmentation and optimization settings are given in
Appendix~\ref{app:impl}.

\subsection{Missing plasma assays}\label{sec:missing}

Missing assays need no imputation. In the prior, a missing marker enters as $z_m = 0$ with $a_m = 0$. To prepare the network for a missing panel, we withhold the whole panel for 10\% of the training samples, and the prior is replaced by the cohort prior, which the regressions predict when all assays are missing. Missing assays are treated in the same way at inference. The model therefore runs on subjects with any subset of the nine assays,
including p-tau217 alone. Section~\ref{sec:res_missing} shows which assays the
model relies on.

\section{Experimental setup}\label{sec:setup}

\subsection{Cohorts}\label{sec:data}

\paragraph{ADNI.}
The development cohort comes from the Alzheimer's Disease Neuroimaging Initiative
(ADNI)~\citep{petersen2010,jack2008}; Appendix~\ref{app:preproc} describes the database, and Table~\ref{tab:cohorts} summarizes both cohorts.
We use 1,132 scan pairs from 952 subjects. Each pair has a T1-weighted MRI, an amyloid PET scan and a plasma sample, with
MRI and plasma taken within 90 days of the PET scan. Three tracers are used:
florbetapir, florbetaben and NAV4694. The plasma panel comes from the University of Pennsylvania ADNI plasma study and has nine measurements. Fujirebio Lumipulse provides p-tau217,
A$\beta$42, A$\beta$40, A$\beta$42/40, p-tau217/A$\beta$42, NfL and GFAP; Quanterix
Simoa provides a second NfL and GFAP~\citep{palmqvist2025,rissin2010}. Every pair has p-tau217, but only
about a third have all nine measurements. We split the pairs by subject into
training, validation and test sets.

\paragraph{A4/LEARN.}
The second cohort combines the A4 study and its companion LEARN study~\citep{sperling2014,sperling2020}.
It differs from ADNI in two ways. First, its participants were cognitively unimpaired
at enrollment. Second, its plasma was measured on other platforms: p-tau217 by Eli
Lilly, and A$\beta$42, A$\beta$40, NfL and GFAP by Roche Elecsys at screening only.
The two Lumipulse NfL and GFAP inputs are never available, and most pairs have
p-tau217 alone. MRI and plasma are taken within 90 days of the PET scan, and the plasma sample is also within 90 days of the MRI. The Roche markers are therefore available only for screening PET scans. Of the 1,645 florbetapir pairs that pass quality control (Appendix~\ref{app:preproc}), we use the 1,473 with a p-tau217 measurement, from 1,006 subjects, and split them by subject into training, validation and test sets. The plasma statistics,
regional regressions and texture template are fitted again on this training set;
nothing is carried over from ADNI.

\paragraph{Preprocessing and amyloid status.}
Both cohorts use the same registration and segmentation. Each pair is registered to the Montreal Neurological Institute (MNI152) space at
1.5~mm resolution ($128\times152\times128$ voxels), and SynthSeg~\citep{billot2023synthseg,billot2023robust} segments
the MRI (Appendix~\ref{app:preproc}). A4/LEARN PET scans are normalized to the cerebellar gray matter (SUVR). ADNI PET scans are distributed intensity-standardized and smoothed to a uniform 6~mm resolution, and we keep this intensity scale. On ADNI, amyloid status is the official status of the UC Berkeley
amyloid PET analysis, which uses tracer-specific thresholds~\citep{landau2012}. On our
SUVR scale, the cortical threshold that best reproduces this status on the training
set is 1.29. We use it to read synthesized scans, and to label the few scans without
an official status (mostly NAV4694). On A4/LEARN, a scan is amyloid-positive when its
true cortical SUVR exceeds 1.32, the threshold that best separates A4 screening scans
from LEARN scans, which were enrolled as amyloid-negative (Appendix~\ref{app:preproc}).

\subsection{Comparison methods}\label{sec:baselines}

We compare with four published synthesis models. Three use MRI only: PASTA, a
conditional diffusion model~\citep{li2024pasta,li2026pasta}, used without its optional clinical inputs; CoCoLIT, a latent diffusion model guided by
a ControlNet on MRI latents~\citep{sargood2026cocolit}; and IL-CLDM, a conditional latent diffusion
model~\citep{ou2025ilcldm}. The fourth, Plasma-CycleGAN, conditions a CycleGAN on plasma A$\beta$42/40~\citep{chen2025plasmacyclegan}. Each model is retrained on every cohort with its published
recipe and chooses its checkpoint on validation. In each cohort, all models use the same training, validation and test pairs. PASTA and Plasma-CycleGAN output intensities scaled to $[0,1]$; we
map them back to SUVR with a constant fitted on the training set. Our ablations
are described with their results in Section~\ref{sec:res_ablation}.

\subsection{Evaluation}\label{sec:eval}

We evaluate each synthesized scan at three levels. Image quality is measured by
PSNR and SSIM inside the brain mask, in SUVR units. Regional amyloid burden is the mean SUVR in the cortex and in the hippocampus, and in three further regions from the amyloid literature. Its agreement with the true scan is
the Spearman rank correlation across test scans, which shows whether a model orders the scans correctly. For amyloid status, the predicted cortical SUVR serves as
a score for the AUC, with DeLong confidence intervals~\citep{delong1988}. We also report sensitivity and specificity at the reading threshold of each cohort, the way a real scan would be read. Differences between models are tested with a
paired bootstrap over test subjects (4,000 resamples). Every model is evaluated at the checkpoint chosen on validation.
Appendix~\ref{app:eval} defines all metrics, and Appendix~\ref{app:metrics} lists
additional results.

\section{Results}\label{sec:results}

\subsection{Comparison with published models}\label{sec:res_main}

\begin{table}[tbp]
\caption{Comparison with published synthesis models on the test sets, at three
levels. Image quality: PSNR and SSIM inside the brain mask, in SUVR units. Regional
burden: Spearman correlation between predicted and true regional SUVR. For each region, $\rho$ is calculated across test scans within each cohort, with one predicted and one measured regional mean per scan. Amyloid
status: AUC for amyloid positivity with a DeLong 95\% confidence interval (CI), against
the official UC Berkeley status on ADNI and true cortical SUVR above 1.32 on
A4/LEARN. Per cohort, the best value in each column is in bold and the second best is underlined.}\label{tab:main}
\begin{tabular*}{\textwidth}{@{\extracolsep\fill}lccccc}
\toprule
 & \multicolumn{2}{c}{Image quality} & \multicolumn{2}{c}{Regional burden ($\rho$)} & Amyloid status \\
\cmidrule{2-3}\cmidrule{4-5}\cmidrule{6-6}
Model & PSNR (dB) & SSIM & Cortex & Hippocampus & AUC [95\% CI] \\
\midrule
\multicolumn{6}{@{}l}{\textit{ADNI} ($n=111$)} \\[2pt]
PASTA           & 20.16 & 0.777 & 0.02     & 0.25 & 0.49 {\footnotesize[0.37, 0.61]} \\
CoCoLIT                    & 19.18 & 0.719 & $-$0.03  & \underline{0.38} & 0.51 {\footnotesize[0.40, 0.63]} \\
IL-CLDM                    & 19.12 & 0.728 & \underline{0.23} & 0.11 & \underline{0.63} {\footnotesize[0.52, 0.74]} \\
Plasma-CycleGAN & \underline{20.70} & \underline{0.791} & 0.13     & 0.27 & 0.55 {\footnotesize[0.45, 0.66]} \\
\arrayrulecolor{black!35}\cmidrule[0.3pt]{1-6}\arrayrulecolor{black}
SUPRA (ours)               & \textbf{23.43} & \textbf{0.836} & \textbf{0.67} & \textbf{0.56} & \textbf{0.91} {\footnotesize[0.85, 0.97]} \\
\midrule
\multicolumn{6}{@{}l}{\textit{A4/LEARN} ($n=143$)} \\[2pt]
PASTA           & 20.43 & 0.810 & $-$0.11  & 0.21 & 0.44 {\footnotesize[0.32, 0.55]} \\
CoCoLIT                    & \underline{21.36} & \underline{0.825} & \underline{0.16} & \underline{0.36} & \underline{0.63} {\footnotesize[0.52, 0.74]} \\
IL-CLDM                    & 16.83 & 0.733 & 0.05 & 0.16 & 0.49 {\footnotesize[0.38, 0.61]} \\
Plasma-CycleGAN & 19.32 & 0.770 & $-$0.01  & 0.13 & 0.51 {\footnotesize[0.40, 0.62]} \\
\arrayrulecolor{black!35}\cmidrule[0.3pt]{1-6}\arrayrulecolor{black}
SUPRA (ours)               & \textbf{22.81} & \textbf{0.864} & \textbf{0.56} & \textbf{0.46} & \textbf{0.82} {\footnotesize[0.72, 0.91]} \\
\botrule
\end{tabular*}
\end{table}

Table~\ref{tab:main} compares our method with the four published models. On ADNI,
our method is best on all five metrics, and every difference to a published model is
significant. The gap in image quality is clear, at 2.7 to 4.3~dB in PSNR. The gap in
amyloid information is much larger. The published models barely track the true
cortical burden, with rank correlations of at most 0.23, and their AUCs for amyloid
status lie between 0.49 and 0.63. Our method reaches a cortical correlation of 0.67
and an AUC of 0.91.

The second cohort shows the same pattern. On A4/LEARN, our method again leads on
every metric. Its differences to the published models are significant on all five
metrics, except for the hippocampal correlation of CoCoLIT ($p=0.15$). The published
models lose most of the cortical information on this cohort, with correlations of at
most 0.16 and AUCs between 0.44 and 0.63. Our method keeps a cortical correlation of 0.56 and an AUC of
0.82.

\begin{figure}[tbp]
\centering
\includegraphics[width=\textwidth]{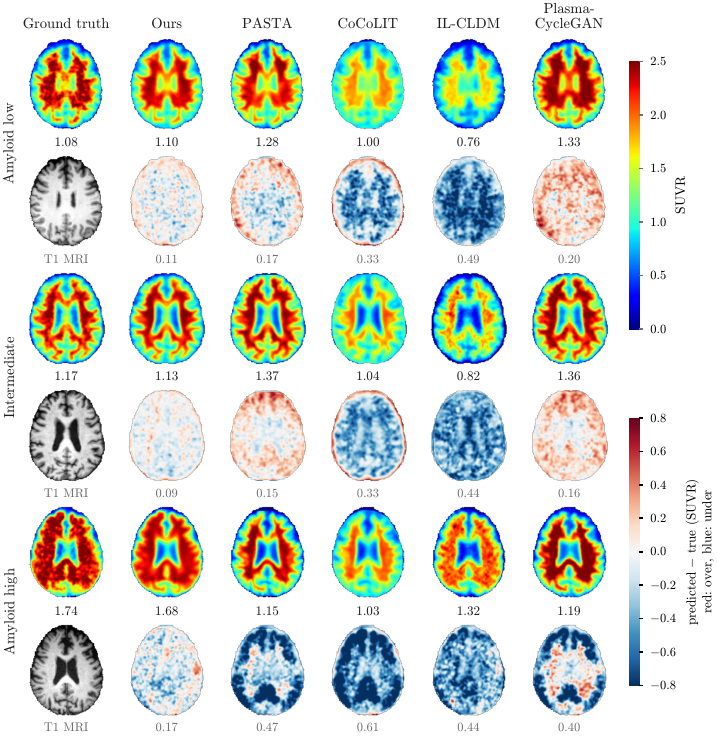}
\caption{Example synthesized scans of three ADNI test subjects with low, intermediate
and high amyloid burden. The subjects come from the three tertiles of true cortical SUVR.
For each subject, the upper row shows the ground truth
and the synthesized PET (absolute SUVR, shared window); the number below is the mean
cortical SUVR. The lower row shows the subject's T1 MRI and the signed error
(predicted $-$ true; red: overestimation, blue: underestimation); the number below
is the mean absolute error inside the brain in this slice.}
\label{fig:qualitative}
\end{figure}

Figure~\ref{fig:qualitative} illustrates these differences on three example
subjects with low, intermediate and high amyloid burden. In these examples, PASTA and
Plasma-CycleGAN overestimate the low and intermediate burden, CoCoLIT and IL-CLDM
underestimate it, and all four underestimate the high burden by 0.4 to 0.7~SUVR.
Over all ADNI test pairs, the median absolute error of the cortical SUVR is
0.08~SUVR for our method and 0.15 to 0.25~SUVR for the published models.

\subsection{Amyloid burden across regions}\label{sec:res_regions}

\begin{table}[tbp]
\caption{Burden agreement in three regions from the amyloid literature: Spearman correlation between
predicted and true regional SUVR on the test sets. Composite: composite cortex; Early: early-accumulation
region (Appendix~\ref{app:eval}). The composite cortex covers only part of the cortex in Table~\ref{tab:main}, so its values differ slightly from the cortical values there. Per cohort, the best value in each column is in bold and the second best is underlined. p-tau217 value: the plasma value used directly as a score.}\label{tab:regions}
\begin{tabular*}{\textwidth}{@{\extracolsep\fill}lcccccc}
\toprule
 & \multicolumn{3}{c}{ADNI} & \multicolumn{3}{c}{A4/LEARN} \\
\cmidrule{2-4}\cmidrule{5-7}
Model & Composite & Early & Striatum & Composite & Early & Striatum \\
\midrule
SUPRA (ours) & \textbf{0.64} & \textbf{0.67} & \textbf{0.56} & \underline{0.59} & \textbf{0.61} & \textbf{0.32} \\
\textit{p-tau217 value} & \underline{0.60} & \underline{0.63} & \underline{0.42} & \textbf{0.64} & \underline{0.58} & \underline{0.28} \\
\cmidrule[0.3pt]{1-7}
PASTA & 0.06 & $-$0.03 & 0.09 & $-$0.14 & $-$0.01 & 0.13 \\
CoCoLIT & $-$0.03 & $-$0.06 & 0.12 & 0.15 & 0.20 & 0.20 \\
IL-CLDM & 0.24 & 0.23 & 0.21 & 0.03 & 0.09 & 0.07 \\
Plasma-CycleGAN & 0.13 & 0.05 & 0.10 & $-$0.01 & 0.04 & 0.06 \\
\botrule
\end{tabular*}
\end{table}

We chose three regions from the amyloid literature to examine regional detail: the
region behind the official amyloid status, and regions affected early and late in the
disease (definitions in Appendix~\ref{app:eval}). The composite cortex is the region
from which UC Berkeley computes the summary SUVR of ADNI amyloid PET and, with a
threshold, the official amyloid status~\citep{landau2015av45methods}. The
early-accumulation region is the part of this composite that two ADNI studies
identified as accumulating amyloid early, namely the precuneus, the posterior cingulate
and the medial orbitofrontal cortex~\citep{palmqvist2017,mattsson2019}, although which
regions come first is not settled~\citep{grothe2017}. The striatum, here the caudate and
the putamen, usually shows amyloid later than the cortex, both at
autopsy~\citep{thal2002} and on PET~\citep{hanseeuw2018,grothe2017}.

Compared with image synthesis or a blood test alone, SUPRA gives more regional detail
(Table~\ref{tab:regions}). The published models produce images with clear brain
structures, but the amyloid levels in these structures barely follow the true ones. On
ADNI, their correlations in the three regions are at most 0.24, 0.23 and 0.21, against
0.64, 0.67 and 0.56 for SUPRA. On A4/LEARN, they are at most 0.15, 0.20 and 0.20,
against 0.59, 0.61 and 0.32. A blood test gives one value per subject, the same for
every region, so it cannot show how amyloid differs between the regions of a subject.
SUPRA gives an SUVR for every structure of the subject's brain, and these values follow
the true burden. Its correlation is higher than that of the plasma p-tau217 value, used directly as a score, in five of
the six columns, most clearly in the ADNI striatum (0.56 against 0.42), and lower in the
A4/LEARN composite cortex (0.59 against 0.64).

\subsection{What each component contributes}\label{sec:res_ablation}

\begin{figure}[tbp]
\centering
\includegraphics[width=\textwidth]{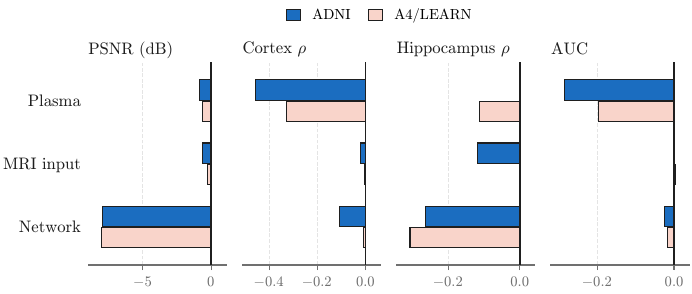}
\caption{Change in each metric when one component is removed from the full model,
on the test sets. Plasma: no plasma input. MRI input: the network receives no
MRI-derived anatomy features. Network: the SUVR prior is used without refinement.
Values are differences to the full model; absolute values are in
Table~\ref{tab:ablation} (Appendix~\ref{app:metrics}).}
\label{fig:ablation}
\end{figure}

Figure~\ref{fig:ablation} shows what each component contributes, by removing one
at a time from the full model. Plasma determines the amyloid information. Without plasma, the network still sees
the MRI and starts from a cohort-average prior, but it can no longer tell subjects
apart by amyloid burden. The cortical correlation drops from 0.67 to 0.21 on ADNI
and from 0.56 to 0.24 on A4/LEARN, and the AUC falls as well (all $p<0.05$). Image quality changes little: PSNR drops by less than 1~dB. The model
therefore produces a realistic image either way, but only plasma makes its amyloid
levels subject-specific.

The refinement network determines the image quality. Without it, the SUVR prior
alone is about 8~dB lower in PSNR on both cohorts. The network also sharpens the
regional pattern: it raises the hippocampal correlation by 0.26 on ADNI and 0.31 on
A4/LEARN. Its effect on cortical burden depends on the cohort. On ADNI it raises the
cortical correlation from 0.56 to 0.67, while on A4/LEARN the prior already ranks
cortical burden as well as the full model.

The MRI input matters least. Without it, the network sees the anatomy only through
the prior, which is painted on the subject's own segmentation. It still keeps most
of the image-quality gain: 7.3 of 8.0~dB on ADNI and 7.8 of 8.0~dB on A4/LEARN. The
MRI input adds the rest, and it improves the hippocampal correlation on ADNI but not
on A4/LEARN. It has no significant effect on cortical
burden or AUC in either cohort. These results match the division of labor
behind the design: plasma sets the amyloid level, the subject's anatomy shapes the
image, and the MRI intensities add a small refinement.

\subsection{Generalization across cohorts}\label{sec:res_transfer}

\begin{table}[tbp]
\caption{Transfer to A4/LEARN ($n=143$, threshold 1.32). The frozen network was
trained on ADNI. In all three settings, the SUVR prior (plasma statistics, regional
regressions and texture template) is fitted on the A4/LEARN training set; the last
row also retrains the network on A4/LEARN. Metrics and marking as in Table~\ref{tab:main}.}\label{tab:transfer}
\begin{tabular*}{\textwidth}{@{\extracolsep\fill}lccccc}
\toprule
 & \multicolumn{2}{c}{Image quality} & \multicolumn{2}{c}{Regional burden ($\rho$)} & Amyloid status \\
\cmidrule{2-3}\cmidrule{4-5}\cmidrule{6-6}
Setting & PSNR (dB) & SSIM & Cortex & Hippocampus & AUC \\
\midrule
SUVR prior only      & 14.81 & 0.415 & \textbf{0.56} & 0.15 & \underline{0.80} \\
ADNI network, frozen & \underline{21.25} & \underline{0.798} & \underline{0.53} & \underline{0.29} & 0.79 \\
\arrayrulecolor{black!35}\cmidrule[0.3pt]{1-6}\arrayrulecolor{black}
Network retrained    & \textbf{22.81} & \textbf{0.864} & \textbf{0.56} & \textbf{0.46} & \textbf{0.82} \\
\botrule
\end{tabular*}
\end{table}

Can a model trained on ADNI serve a cohort whose plasma was measured on other
platforms? The ADNI regressions cannot be applied directly, because the assays
differ. We therefore refit only the SUVR prior on the A4/LEARN training set: the
plasma statistics, the regional regressions and the texture template. These are
closed-form steps that take minutes and need no network training. The network
trained on ADNI is kept frozen. Table~\ref{tab:transfer} compares this setting with
the prior alone and with a model retrained on A4/LEARN.

The refitted prior carries the amyloid information to the new cohort. With the
frozen network, the cortical correlation and the AUC do not differ significantly
from those of the retrained model (Table~\ref{tab:transfer}). The frozen network
also keeps most of its benefit for image quality over the prior alone. The remaining
gap lies in image detail: the retrained model is better by 1.6~dB in PSNR ($p<0.001$) and in the hippocampal correlation ($p<0.05$). Part of this gap may reflect resolution: the frozen network was trained on ADNI images smoothed to 6~mm, and the A4/LEARN scans are not smoothed. A new
cohort can thus be served by refitting the prior alone, and retraining the network
improves the image further.

\subsection{Which plasma markers matter}\label{sec:res_missing}

\begin{table}[tbp]
\caption{Contribution of each plasma marker group to the cortical SUVR, on the
training sets. $R^2$ is the leave-one-out explained variance of the regional
regression for log cortical SUVR, fitted with the group alone; $\Delta R^2$ is the
loss when the group is removed from the full panel. p-tau217: p-tau217 and
p-tau217/A$\beta$42. A$\beta$: A$\beta$42, A$\beta$40 and A$\beta$42/40. NfL/GFAP:
both platforms. The largest value in each column is in bold and the second largest is
underlined.}\label{tab:markers}
\begin{tabular*}{\textwidth}{@{\extracolsep\fill}lcccc}
\toprule
 & \multicolumn{2}{c}{ADNI} & \multicolumn{2}{c}{A4/LEARN} \\
\cmidrule{2-3}\cmidrule{4-5}
Marker group & $R^2$ alone & $\Delta R^2$ when removed & $R^2$ alone & $\Delta R^2$ when removed \\
\midrule
All markers  & \textbf{0.51} & --   & \textbf{0.30} & -- \\
\arrayrulecolor{black!35}\cmidrule[0.3pt]{1-5}\arrayrulecolor{black}
p-tau217     & \underline{0.48} & \textbf{0.25} & \textbf{0.30} & \textbf{0.20} \\
A$\beta$     & 0.18 & \underline{0.02} & 0.08 & 0.00 \\
NfL/GFAP     & 0.13 & 0.01 & \underline{0.10} & \underline{0.01} \\
\botrule
\end{tabular*}
\end{table}

Because the prior is a regression on the plasma markers, we can ask directly which
markers carry the amyloid information. We group the nine measurements into
p-tau217, A$\beta$ and NfL/GFAP, and fit the regression for cortical SUVR on the
training set with each group alone and with the full panel minus that group
(Table~\ref{tab:markers}).

p-tau217 carries almost all of the information (Table~\ref{tab:markers}). On ADNI it
explains 0.48 of the 0.51 variance that the full panel explains, and on A4/LEARN it
explains all of it. Removing it costs far more than removing the A$\beta$ or
NfL/GFAP markers. These markers correlate with amyloid burden on their own, but they
add little once p-tau217 is known. Plasma explains little of the hippocampal SUVR
($R^2 \le 0.10$), which agrees with the ablation in
Section~\ref{sec:res_ablation}: the hippocampal pattern comes from the images.

The full model behaves accordingly when assays are withheld at inference. With
p-tau217 as its only plasma input, it reaches a cortical correlation of 0.59 on ADNI,
compared with 0.67 for the full panel (difference not significant), and the same
0.56 on A4/LEARN. Without any plasma input at inference, the correlation falls to 0.32 and $-0.01$ (the ablation in Section~\ref{sec:res_ablation} is instead trained without plasma). A single p-tau217 measurement is thus enough to use the model.

\subsection{Amyloid positivity as a decision}\label{sec:res_positivity}

\begin{figure}[tbp]
\centering
\includegraphics[width=\textwidth]{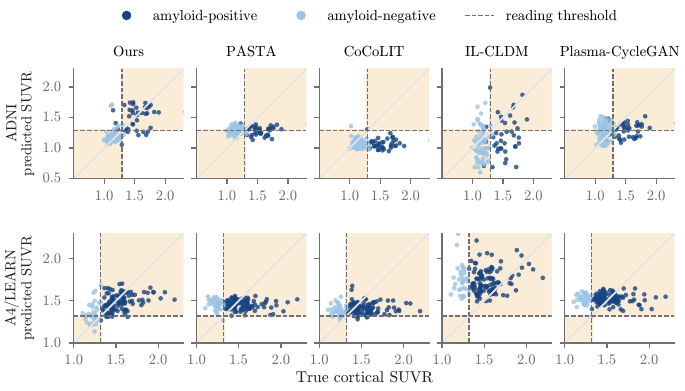}
\caption{Predicted against true mean cortical SUVR for every test scan. Dashed
lines: reading threshold (1.29 for ADNI, 1.32 for A4/LEARN); shaded quadrants: true
and predicted SUVR on the same side of it. Colors give the true amyloid status
(dark blue positive, light blue negative): the official UC Berkeley status on ADNI and
true SUVR above 1.32 on A4/LEARN. }
\label{fig:positivity_scatter}
\end{figure}

A real amyloid PET scan is read by comparing its SUVR with a fixed threshold. A
synthesized scan can be read the same way only if two conditions hold: its values
must be in absolute SUVR, and they must differ between subjects as the true values
do. Figure~\ref{fig:positivity_scatter} shows that the published models fail the
second condition. Three of them predict nearly the same cortical SUVR for every
subject, with a standard deviation of 0.06 to 0.13 on ADNI against a true 0.24.
IL-CLDM varies more, but largely independently of the true burden. As a result,
their decisions are close to chance. On A4/LEARN, all four call almost every subject
positive (specificity at most 0.14), and on ADNI, CoCoLIT calls almost every subject
negative (sensitivity 0.02). Our predictions instead follow the true burden, with AUCs of 0.91 on ADNI and 0.82 on A4/LEARN, and can be read against a threshold, provided that it is calibrated on data from the same cohort (Appendix~\ref{app:metrics}).

\section{Discussion}\label{sec:discussion}

\paragraph{Amyloid information comes from plasma.}
Our results show that a synthesized amyloid PET scan can carry subject-specific
amyloid information when the synthesis uses a blood test. The published models
reproduce the appearance of an amyloid PET scan, but they barely track cortical
amyloid burden, and they separate amyloid-positive from amyloid-negative subjects
poorly. This holds for the three models that see only the MRI,
and also for Plasma-CycleGAN, which uses plasma A$\beta$42/40. In our model, the ablations locate
the source of the amyloid information. Without plasma, image quality stays almost
the same, but the cortical correlation falls to less than half of its value in
both cohorts.
The regression analysis agrees: p-tau217 explains almost all of the cortical SUVR
that the plasma panel can explain, while plasma explains little of the hippocampal
SUVR, whose pattern comes from the images.

\paragraph{How the inputs share the work.}
The ablations support the division of labor behind the design. Plasma sets the
amyloid level, the refinement network turns the prior into a realistic image, and
the subject's anatomy places the amyloid on the right structures. Most of this
anatomical information already reaches the network through the prior, which is
painted on the subject's own segmentation. The MRI intensities add a smaller
refinement: about 0.6~dB on ADNI and 0.2~dB on A4/LEARN, and a better hippocampal
pattern on ADNI. A practical consequence is that the two inputs complement each other. A blood test indicates the amyloid level of a subject but gives no image, and an MRI scan gives a realistic image but says little about the amyloid level. The synthesis combines both.

\paragraph{Regional amyloid information.}
The synthesized scans also carry amyloid information beyond the cortical average. In both
cohorts, SUPRA follows the true burden across subjects in the composite cortex and the
early-accumulation region, and to a lesser degree in the striatum, whereas the published
models barely do (Table~\ref{tab:regions}). The early-accumulation region and the striatum
matter because amyloid pathology is staged by the order in which it reaches brain
regions~\citep{thal2002,grothe2017,mattsson2019,hanseeuw2018}. A synthesized scan thus adds
to the blood test an estimate in absolute SUVR for each region.

\paragraph{Reading synthesized scans against a threshold.}
Because the output is in absolute SUVR, a synthesized scan can be read like a real
one, against a positivity threshold. On ADNI, this already gives a balanced
decision. Our predictions are, however, pulled toward the cohort mean, most clearly
on A4/LEARN, where amyloid-negative subjects are predicted slightly too high and the specificity drops (Table~\ref{tab:status}). A threshold recalibrated on validation data from the same
cohort restores a more useful balance. Any use of synthesized scans for decisions should therefore include a calibration step for the target cohort. Such a calibration needs a reference sample with true PET from the target population, so a threshold that holds across populations remains an open problem. Synthesized scans are
not a replacement for amyloid PET. They could, however, provide regional amyloid
estimates where PET is not available, for example in large cohorts that collect MRI and blood samples but no amyloid PET. These uses
would need prospective evaluation.

\paragraph{Evaluating synthesized amyloid PET.}
Image-similarity metrics alone can be misleading for amyloid PET synthesis. In our
comparison, the published models reach reasonable PSNR and SSIM values while their
cortical SUVR hardly follows the true burden (Table~\ref{tab:calib}). Studies of
amyloid PET synthesis should therefore also report regional burden agreement and
amyloid status, in absolute SUVR units.

\paragraph{Strengths.}
We evaluated the method in two cohorts that differ in cognitive profile and plasma assay platforms, and we tested transfer between them by refitting only the prior. The architecture is transparent, and the ablations locate the contribution of each input. The method performs better than four published synthesis models in both cohorts.

\paragraph{Limitations.}
Our study has several limitations. Both cohorts are research cohorts with selected
participants, and the A4/LEARN participants were cognitively unimpaired. The
performance in memory-clinic populations remains to be tested. Our predictions are
compressed toward the cohort mean, so high burden tends to be underestimated and low
burden overestimated. The comparison methods were trained with their published recipes and
not tuned further; tuning might improve their image quality.

\paragraph{Future work.}
Several extensions follow naturally. The same design could be applied to tau PET,
where plasma p-tau217 may also be informative, and to longitudinal data, to synthesize
changes in amyloid burden over time. A per-subject uncertainty estimate would help users decide when a synthesized scan can be trusted. Further work could test the method in cohorts earlier in the disease course and examine whether synthesized scans help predict cognitive decline. Finally, evaluation in
clinical cohorts with routine MRI and clinical plasma assays would show how well the
method works outside research studies.

\section{Conclusion}\label{sec:conclusion}

We presented SUPRA, a method that synthesizes amyloid PET from structural MRI and
plasma biomarkers. It turns the plasma panel into an explicit regional SUVR prior,
painted on the subject's own anatomy, and refines this prior with an anatomy-guided
residual network. The output is in absolute SUVR, and the model accepts incomplete
plasma panels, down to p-tau217 alone. On ADNI and on the second cohort, A4/LEARN,
the synthesized scans reach higher image quality than the published synthesis models, and they track the true amyloid burden across brain regions and separate amyloid-positive from amyloid-negative subjects far better. Our ablations show that the amyloid information comes from plasma,
while the subject's anatomy shapes the image. In our experiments, removing the blood test removed most of the amyloid information. Combining the two inputs gives images that are realistic and carry subject-specific amyloid information in each region, which a single blood value cannot show.

\section*{Abbreviations}

A4: Anti-Amyloid Treatment in Asymptomatic Alzheimer's; A$\beta$: Amyloid-$\beta$; AdaGN: Adaptive group normalization; ADNI: Alzheimer's Disease Neuroimaging Initiative; ANTs: Advanced Normalization Tools; ATRI: Alzheimer's Therapeutic Research Institute; AUC: Area under the receiver operating characteristic curve; BFDN: Bridge Funding for Disrupted Neurodegenerative Research; CI: Confidence interval; CN: Cognitively normal; CSF: Cerebrospinal fluid; EMA: Exponential moving average; FDG: Fluorodeoxyglucose; FN: False negative; FNIH: Foundation for the National Institutes of Health; FP: False positive; GFAP: Glial fibrillary acidic protein; GN: Group normalization; GRIP: Global Research \& Imaging Platform; LEARN: Longitudinal Evaluation of Amyloid Risk and Neurodegeneration; LoA: Limits of agreement; MAE: Mean absolute error; MCI: Mild cognitive impairment; MNI: Montreal Neurological Institute; MRI: Magnetic resonance imaging; NfL: Neurofilament light chain; p-tau217: Phosphorylated tau 217; PET: Positron emission tomography; PSNR: Peak signal-to-noise ratio; ROI: Region of interest; SD: Standard deviation; SiLU: Sigmoid linear unit; SSIM: Structural similarity index; SUPRA: SUVR prior-guided amyloid synthesis; SUVR: Standardized uptake value ratio; TN: True negative; TP: True positive; WM: White matter.

\section*{Declarations}

\subsection*{Availability of data and materials}
The data that support the findings of this study are available from the ADNI database~\citep{adnidata} and from the A4 and LEARN data platform~\citep{a4data}, but restrictions apply to the availability of these data, which were used under data use agreements for the current study, and so are not publicly available. Qualified researchers can apply for access to both data sources. The code will be made publicly available upon publication.

\subsection*{Competing interests}
The authors declare that they have no competing interests.

\subsection*{Funding}
This work was supported by the Alzheimer's Association and The Michael J. Fox Foundation for Parkinson's Research through the Bridge Funding for Disrupted Neurodegenerative Research (BFDN) Grant Program (Grant No. 26BFDN-1609491C). The funders had no role in the study design, data analysis, decision to publish, or preparation of the manuscript.

\subsection*{Authors' contributions}
HL performed the experiments and wrote the manuscript. CY conceived the study and supervised the work. FD, ZW, RT, FL, BB, MSB and AW provided guidance and commented on the manuscript. All authors read and approved the final manuscript.

\subsection*{Acknowledgements}
Data collection and sharing for the Alzheimer's Disease Neuroimaging Initiative (ADNI) is funded by the National Institute on Aging (National Institutes of Health Grant U19AG024904). The grantee organization is the Northern California Institute for Research and Education. In the past, ADNI has also received funding from the National Institute of Biomedical Imaging and Bioengineering, the Canadian Institutes of Health Research, and private sector contributions through the Foundation for the National Institutes of Health (FNIH).

The A4 Study was a secondary prevention trial in preclinical Alzheimer's disease, aiming to slow cognitive decline associated with brain amyloid accumulation in clinically normal older individuals. The A4 Study was funded by a public-private-philanthropic partnership, including funding from the National Institutes of Health-National Institute on Aging, Eli Lilly and Company, Alzheimer's Association, Accelerating Medicines Partnership, GHR Foundation, an anonymous foundation, and additional private donors, with in-kind support from Avid Radiopharmaceuticals, Cogstate, Albert Einstein College of Medicine and the Foundation for Neurologic Diseases. The companion observational Longitudinal Evaluation of Amyloid Risk and Neurodegeneration (LEARN) Study was funded by the Alzheimer's Association and GHR Foundation. The A4 and LEARN Studies were led by Dr. Reisa Sperling at Brigham and Women's Hospital, Harvard Medical School, and Dr. Paul Aisen at the Alzheimer's Therapeutic Research Institute (ATRI) at the University of Southern California. The A4 and LEARN Studies were coordinated by ATRI at the University of Southern California, and the data are made available under the auspices of Alzheimer's Clinical Trial Consortium through the Global Research \& Imaging Platform (GRIP). The complete A4 Study Team list is available on: \url{https://www.actcinfo.org/a4-study-team-lists/}. We would like to acknowledge the dedication of the study participants and their study partners who made the A4 and LEARN Studies possible.

\bibliographystyle{iclr2027_conference}
\bibliography{sn-bibliography}

\clearpage
\begin{appendices}
\counterwithin*{table}{section}\counterwithin*{figure}{section}
\renewcommand{\thetable}{\Alph{section}\arabic{table}}
\renewcommand{\thefigure}{\Alph{section}\arabic{figure}}
\renewcommand{\theHtable}{app.\thetable}\renewcommand{\theHfigure}{app.\thefigure}

\section{Cohorts, preprocessing and amyloid thresholds}\label{app:preproc}

\begin{table}[htbp]
\caption{Cohort characteristics per data split. A scan pair combines a T1-weighted MRI,
an amyloid PET scan and a plasma sample; the splits are by subject. Age and sex are
counted per pair, with age (mean $\pm$ standard deviation) at the PET scan. The ADNI diagnosis is taken from the visit
closest to the PET scan (within 180 days); 7 training and 2 test pairs have no such
visit. Amyloid status as in Section~\ref{sec:data}: official UC Berkeley status on
ADNI, true cortical SUVR above 1.32 on A4/LEARN. A4/LEARN participants were
cognitively unimpaired at enrollment. The A4/LEARN
training set also contains one pair from an A4 screen-failure participant. CN: cognitively normal; MCI: mild cognitive impairment.}\label{tab:cohorts}
{\setlength{\tabcolsep}{2pt}%
\begin{tabular*}{\textwidth}{@{\extracolsep\fill}lcccccc}
\toprule
 & \multicolumn{3}{c}{ADNI} & \multicolumn{3}{c}{A4/LEARN} \\
\cmidrule{2-4}\cmidrule{5-7}
 & Train & Validation & Test & Train & Validation & Test \\
\midrule
Scan pairs              & 906 & 115 & 111 & 1195 & 135 & 143 \\
Subjects                & 762 & 95  & 95  & 810   & 99  & 97  \\
Age (years)             & 73.0 $\pm$ 8.1 & 72.6 $\pm$ 8.0 & 72.7 $\pm$ 9.4 & 73.7 $\pm$ 5.2 & 73.7 $\pm$ 4.9 & 73.8 $\pm$ 5.1 \\
Female (\%)             & 53.9 & 66.1 & 53.2 & 60.1 & 48.1 & 62.9 \\
CN/MCI/dementia         & 506/293/100 & 71/34/10 & 70/31/8 & \multicolumn{3}{c}{all unimpaired} \\
Amyloid-positive (\%)  & 44.9 & 39.1 & 37.8 & 77.4 & 76.3 & 75.5 \\[2pt]
\multicolumn{7}{@{}l}{\textit{PET tracer (pairs)}} \\
\quad Florbetapir       & 438 & 54  & 61  & 1195 & 135 & 143 \\
\quad Florbetaben       & 426 & 57  & 46  & --    & --  & --  \\
\quad NAV4694           & 42  & 4   & 4   & --    & --  & --  \\[2pt]
\multicolumn{7}{@{}l}{\textit{Study (pairs)}} \\
\quad A4                & --  & --  & --  & 1104 & 124 & 133 \\
\quad LEARN             & --  & --  & --  & 90    & 11  & 10  \\[2pt]
\multicolumn{7}{@{}l}{\textit{Plasma panel (\% of pairs)}} \\
\quad p-tau217 measured & 100 & 100 & 100 & 100   & 100 & 100 \\
\quad All nine measurements & 33 & 30 & 42 & 0  & 0   & 0   \\
\quad p-tau217 only     & 0   & 0   & 1   & 72    & 75  & 69  \\
\botrule
\end{tabular*}}
\end{table}

Data used in the preparation of this article were obtained from the ADNI database (adni.loni.usc.edu) in May 2026. ADNI was launched in 2003 as a public-private partnership, led by Principal Investigator Michael W. Weiner, MD. Its original goal was to test whether serial MRI, PET, other biological markers, and clinical and neuropsychological assessment can be combined to measure the progression of mild cognitive impairment and early Alzheimer's disease.

T1 scans are skull-stripped with SynthStrip~\citep{hoopes2022}. The PET scan is
skull-stripped in the same way, registered rigidly to the T1 scan, and the T1 scan
is registered affinely to the MNI152 template with Advanced Normalization Tools (ANTs)~\citep{avants2011}. The two transforms
are composed, so the PET scan is interpolated only once. T1 intensities are
normalized with WhiteStripe~\citep{shinohara2014}. Both images are resampled to 1.5~mm and
padded to $128\times152\times128$ voxels. SynthSeg~\citep{billot2023synthseg,billot2023robust} (robust mode)
segments the T1 scan in this space. For ADNI we start from the intensity-standardized PET images distributed by ADNI (``Coreg, Avg, Std Img and Vox Siz, Uniform 6mm Res''); their mean value in the cerebellar gray matter is $0.98\pm0.06$. For A4/LEARN we realign and average the dynamic frames ourselves, without smoothing, and normalize each scan to its cerebellar gray matter (SynthSeg labels).
Quality control on A4/LEARN (brain volumes, registration and reference-region
coverage) keeps 1,645 of 1,666 pairs.

We set the thresholds from the data. On the ADNI training set, the cortical SUVR that best reproduces the official status is 1.29, with an agreement of 93\% (1.26 for florbetapir and 1.29 for florbetaben alone). On A4/LEARN, true cortical SUVR separates A4 screening scans from LEARN scans with an AUC of 0.93, and the best cut is 1.32. The threshold of 1.11 often used for florbetapir with a whole-cerebellum reference~\citep{landau2012} does not fit our SUVR scale: on A4/LEARN it labels 90\% of LEARN scans positive, although LEARN enrolled amyloid-negative participants, and on the ADNI training set it agrees with the official UC Berkeley status for only 66\% of scans.

\section{Implementation details}\label{app:impl}

The anatomy encoder has five stages with 16, 32, 64, 96 and 128 channels. Each stage
applies two $3\times3\times3$ convolutions with group normalization and sigmoid linear unit (SiLU) activation, and consecutive stages are separated by $2\times$ average pooling. The
U-Net has five encoder levels with 24, 48, 96, 192 and 256 channels, a bottleneck
block and four decoder levels, joined by skip connections. Each level is one residual
block with two $3\times3\times3$ convolutions. Downsampling uses average pooling, and
upsampling uses trilinear interpolation. The U-Net input has 18 channels: the prior as
two identical channels and the 16 channels of $\mathbf{F}^{(1)}$. The model has 16.2 million
parameters, 1.4 million of them in the anatomy encoder.

During training, all input and target volumes, including the prior, receive the same
random rotation (up to $6^\circ$), scaling (up to 6\%), translation (up to 4\% of the
field of view) and left--right flip. MRI intensities are further varied by a random
gamma change, a smooth bias field and Gaussian noise. PET values are never rescaled,
so the SUVR scale of the target is preserved. The SSIM term uses a fixed data range
of 3.244 SUVR.

We train with AdamW (learning rate $2\times10^{-4}$, weight decay $10^{-5}$, momentum
parameters $(0.9, 0.99)$), a batch size of 4, gradient clipping at 1.0 and mixed
precision, for 70 epochs. The learning rate follows a slow cosine decay and ends at
$1.86\times10^{-4}$. We keep an exponential moving average (EMA) of the weights with
decay 0.999 and use the EMA weights for validation and testing. Every five epochs,
the EMA model is evaluated on the validation split, and we keep the checkpoint with
the best value of a fixed composite score. This score combines masked and unmasked
PSNR and SSIM, cortical and hippocampal burden agreement, the positivity AUC, and
penalties on the standard deviation (SD) ratio and the mean bias. After training, the best and the last checkpoint
are compared on validation with the same score. Inference is a single forward pass; negative
output values are set to zero.

\section{Evaluation metrics}\label{app:eval}

All metrics are computed in absolute SUVR on the $128\times152\times128$ grid. Let $Y$
be the true and $\hat{Y}$ the synthesized PET of a test pair, and let $\Omega$ be the
set of evaluated voxels: the brain mask $M$, or the whole volume for the unmasked
metrics in Table~\ref{tab:voxel}.

\paragraph{Voxel-level metrics.}
The mean absolute error (MAE) and PSNR are
\begin{equation}
\mathrm{MAE} = \frac{1}{|\Omega|}\sum_{v\in\Omega} \big|\hat{Y}(v) - Y(v)\big|, \qquad
\mathrm{PSNR} = 10\log_{10}\frac{L^2}{\frac{1}{|\Omega|}\sum_{v\in\Omega} \big(\hat{Y}(v) - Y(v)\big)^2},
\end{equation}
with a fixed data range $L = 3.244$~SUVR for all models and both cohorts. The
structural similarity~\citep{wang2004} is computed with a Gaussian window (standard
deviation of 1.5 voxels):
\begin{equation}
\mathrm{SSIM}(v) = \frac{(2\mu_{\hat{Y}}\mu_{Y} + c_1)(2\sigma_{\hat{Y}Y} + c_2)}
{(\mu_{\hat{Y}}^2 + \mu_{Y}^2 + c_1)(\sigma_{\hat{Y}}^2 + \sigma_{Y}^2 + c_2)},
\end{equation}
where $\mu$, $\sigma^2$ and $\sigma_{\hat{Y}Y}$ are the local means, variances and
covariance in the window around $v$, $c_1 = (0.01L)^2$ and $c_2 = (0.03L)^2$. The SSIM
map is computed over the whole volume and then averaged over $\Omega$, so windows at
the brain edge keep their true neighborhood. All three metrics are averaged over the
test pairs.

\paragraph{Regional burden.}
The burden of a region $R$ is the mean SUVR inside it. The cortex is given by the
SynthSeg labels 3 and 42 and the hippocampus by the labels 17 and 53, both
intersected with the brain mask. The same label map, from the subject's MRI, is used
for the true and the synthesized scan. Let $b_i$ and $\hat{b}_i$ be the true and the
predicted burden of test pair $i$. The burden agreement is the Spearman rank
correlation $\rho$ between $\hat{b}_i$ and $b_i$ over the test pairs. It depends only
on the ranks, so a shift or rescaling of all predictions does not change it.

\paragraph{Regions from the amyloid literature.}
We also report the burden agreement in three further regions (Section~\ref{sec:res_regions}).
The composite cortex consists of 20 bilateral
Desikan-Killiany parcels: nine frontal (caudal and rostral middle frontal, superior
frontal, lateral and medial orbitofrontal, pars opercularis, pars triangularis, pars
orbitalis and frontal pole), four cingulate (caudal and rostral anterior cingulate,
posterior cingulate and isthmus), four parietal (inferior parietal, superior parietal,
supramarginal and precuneus) and three temporal (superior, middle and inferior temporal). The parcels come from SynthSeg~\citep{billot2023synthseg,billot2023robust}, run on the T1 scan in MNI space with FreeSurfer (\texttt{mri\_synthseg --robust --parc}). The definition follows the current official UC Berkeley table, which, unlike
the 2015 description~\citep{landau2015av45methods}, includes the inferior temporal gyrus: the volumes of these 20 parcels
sum to the summary volume of the table on all 4,377 scans we checked. We apply the
same definition to A4/LEARN. The early-accumulation region consists of the precuneus,
the posterior cingulate with the isthmus and the medial orbitofrontal cortex, the
regions shared by the two ADNI studies~\citep{palmqvist2017,mattsson2019}. These studies
use longitudinal rates, whereas we evaluate single scans. The striatum consists of the
caudate and the putamen (SynthSeg labels 11, 50, 12 and 51), without the nucleus
accumbens.

\paragraph{Amyloid status.}
The predicted cortical burden $\hat{b}_i$ serves as the score for amyloid positivity.
The AUC is the probability that a randomly chosen positive scan has a higher score than a randomly chosen negative scan; its 95\% confidence interval follows \citet{delong1988}. At a threshold $c$, a scan is called positive when
$\hat{b}_i > c$, and
\begin{equation}
\text{sensitivity} = \frac{\mathrm{TP}}{\mathrm{TP} + \mathrm{FN}}, \qquad
\text{specificity} = \frac{\mathrm{TN}}{\mathrm{TN} + \mathrm{FP}},
\end{equation}
where TP, FN, TN and FP count the true positives, false negatives, true negatives and
false positives. We use the reading threshold of each cohort and, for our models, the
recalibrated threshold. The latter maximizes the Youden index,
$J = \text{sensitivity} + \text{specificity} - 1$, on the validation predictions.

\paragraph{Calibration.}
With the cortical error $d_i = \hat{b}_i - b_i$, Table~\ref{tab:calib} reports the
least-squares slope of $\hat{b}$ on $b$, the SD ratio
$\operatorname{sd}(\hat{b}) / \operatorname{sd}(b)$, the mean bias $\bar{d}$, and the
half-width of the 95\% limits of agreement, $1.96\,\operatorname{sd}(d)$.
Section~\ref{sec:res_main} also reports the median of $|d_i|$.

\paragraph{Statistical comparison.}
Two models are compared on the same test pairs by a paired bootstrap. We draw 4,000 resamples of the test subjects with replacement, keeping all pairs of each subject, and compute each metric for both models
on every resample. Let $\Delta^{*}$ be the difference between the two models on a
resample. The two-sided $p$-value is
$p = 2\min\{P(\Delta^{*} \le 0),\, P(\Delta^{*} \ge 0)\}$, capped at 1. Resampling whole subjects keeps the pairing between the models, and the rank statistics ($\rho$ and
AUC) are recomputed on every resample.

\section{Additional metrics}\label{app:metrics}

\begin{table}[htbp]
\caption{Ablation of the main components on the test sets. Plasma: the plasma panel
enters the model (through the SUVR prior). MRI: the network
receives the MRI-derived anatomy features. When this input is omitted, the prior still uses the subject's MRI segmentation. Network: the refinement network is
applied; without it, the SUVR prior is the output. Metrics and marking as in Table~\ref{tab:main}; the same values are shown as differences in Fig.~\ref{fig:ablation}.}\label{tab:ablation}
{\setlength{\tabcolsep}{2.5pt}%
\begin{tabular*}{\textwidth}{@{\extracolsep\fill}lcccccccc}
\toprule
 & \multicolumn{3}{c}{Components} & \multicolumn{2}{c}{Image quality} & \multicolumn{2}{c}{Regional burden ($\rho$)} & \begin{tabular}[b]{@{}c@{}}Amyloid\\status\end{tabular} \\
\cmidrule{2-4}\cmidrule{5-6}\cmidrule{7-8}\cmidrule{9-9}
Model & Plasma & MRI & Network & PSNR (dB) & SSIM & Cortex & Hippocampus & AUC \\
\midrule
\multicolumn{9}{@{}l}{\textit{ADNI} ($n=111$)} \\[2pt]
SUPRA (ours)      & \checkmark & \checkmark & \checkmark & \textbf{23.43} & \textbf{0.836} & \textbf{0.67} & \textbf{0.56} & \textbf{0.91} \\
No plasma          & --         & \checkmark & \checkmark & 22.56 & \underline{0.824} & 0.21 & \textbf{0.56} & 0.62 \\
\shortstack[l]{No MRI input to the\\refinement network} & \checkmark & --         & \checkmark & \underline{22.83} & 0.809 & \underline{0.65} & \underline{0.44} & \textbf{0.91} \\
SUVR prior only    & \checkmark & --         & --         & 15.48 & 0.414 & 0.56 & 0.30 & \underline{0.88} \\
\midrule
\multicolumn{9}{@{}l}{\textit{A4/LEARN} ($n=143$)} \\[2pt]
SUPRA (ours)      & \checkmark & \checkmark & \checkmark & \textbf{22.81} & \textbf{0.864} & \textbf{0.56} & \textbf{0.46} & \textbf{0.82} \\
No plasma          & --         & \checkmark & \checkmark & 22.17 & \underline{0.857} & \underline{0.24} & \underline{0.35} & 0.62 \\
\shortstack[l]{No MRI input to the\\refinement network} & \checkmark & --         & \checkmark & \underline{22.56} & 0.851 & \textbf{0.56} & \textbf{0.46} & \textbf{0.82} \\
SUVR prior only    & \checkmark & --         & --         & 14.81 & 0.415 & \textbf{0.56} & 0.15 & \underline{0.80} \\
\botrule
\end{tabular*}}
\end{table}

Tables~\ref{tab:voxel}--\ref{tab:status} list further metrics for the published
models, the SUVR prior and our model, on the same test pairs as Table~\ref{tab:main}.
Table~\ref{tab:status} also includes the two ablation models of
Table~\ref{tab:ablation}. Three points stand out.

First, PSNR and SSIM over the whole volume are higher than inside the brain mask for all models, because the background is easy to predict; the exception is the PSNR of Plasma-CycleGAN. Our
model remains best on both versions of each metric, in both cohorts.

Second, the cortical SUVR of the published models hardly follows the true burden. Their slopes lie between $-0.02$ and 0.08, except for IL-CLDM on ADNI (0.33). PASTA, CoCoLIT and Plasma-CycleGAN also vary little between subjects (SD ratios of 0.25 to 0.53), while IL-CLDM varies more (0.85 and 1.17). Our predictions follow the
true values, with a slope of 0.57 on ADNI and 0.28 on A4/LEARN and a mean bias of at
most 0.03 SUVR. They are still pulled toward the cohort mean, most clearly on
A4/LEARN, where the SD ratio is 0.54.

Third, on A4/LEARN this pull explains the low specificity at the reading threshold
(Table~\ref{tab:status}). Amyloid-negative subjects are predicted slightly
too high and cross the threshold. The recalibrated threshold, chosen on the
validation set, raises the specificity of our model on this cohort to 0.71, at a sensitivity of 0.75.

As a reference, we used the plasma p-tau217 value itself as a score on the same test pairs. It gives AUCs of 0.89 on ADNI and 0.80 on A4/LEARN and cortical correlations of 0.58 and 0.61.

\begin{table}[htbp]
\caption{Voxel-level metrics on the test sets: mean absolute error (MAE, SUVR) inside the
brain mask, and PSNR and SSIM inside the brain mask and over the whole volume. Per cohort, the best value in each column
(lowest MAE, highest PSNR and SSIM) is in bold and the second best is underlined.}\label{tab:voxel}
\begin{tabular*}{\textwidth}{@{\extracolsep\fill}lccccc}
\toprule
 & & \multicolumn{2}{c}{Brain mask} & \multicolumn{2}{c}{Whole volume} \\
\cmidrule{3-4}\cmidrule{5-6}
Model & MAE & PSNR (dB) & SSIM & PSNR (dB) & SSIM \\
\midrule
\multicolumn{6}{@{}l}{\textit{ADNI} ($n=111$, threshold 1.29)} \\[2pt]
PASTA              & 0.252 & 20.16 & 0.777 & \underline{25.91} & \underline{0.897} \\
CoCoLIT            & 0.309 & 19.18 & 0.719 & 24.93 & 0.884 \\
IL-CLDM            & 0.320 & 19.12 & 0.728 & 24.86 & 0.884 \\
Plasma-CycleGAN    & \underline{0.249} & \underline{20.70} & \underline{0.791} & 19.17 & 0.853 \\
\arrayrulecolor{black!35}\cmidrule[0.3pt]{1-6}\arrayrulecolor{black}
SUVR prior only    & 0.382 & 15.48 & 0.414 & 21.36 & 0.760 \\
SUPRA (ours)       & \textbf{0.183} & \textbf{23.43} & \textbf{0.836} & \textbf{28.97} & \textbf{0.925} \\
\midrule
\multicolumn{6}{@{}l}{\textit{A4/LEARN} ($n=143$, threshold 1.32)} \\[2pt]
PASTA              & 0.245 & 20.43 & 0.810 & 26.18 & 0.908 \\
CoCoLIT            & \underline{0.234} & \underline{21.36} & \underline{0.825} & \underline{27.04} & \underline{0.916} \\
IL-CLDM            & 0.400 & 16.83 & 0.733 & 22.59 & 0.849 \\
Plasma-CycleGAN    & 0.288 & 19.32 & 0.770 & 17.00 & 0.834 \\
\arrayrulecolor{black!35}\cmidrule[0.3pt]{1-6}\arrayrulecolor{black}
SUVR prior only    & 0.390 & 14.81 & 0.415 & 20.70 & 0.750 \\
SUPRA (ours)       & \textbf{0.195} & \textbf{22.81} & \textbf{0.864} & \textbf{28.41} & \textbf{0.930} \\
\botrule
\end{tabular*}
\end{table}

\begin{table}[htbp]
\caption{Calibration of the predicted cortical SUVR against the true cortical SUVR on the
test sets. Slope: least-squares slope of predicted on true values (1 is ideal). SD
ratio: standard deviation of the predicted values divided by that of the true values (1
is ideal; below 1, predictions are pulled toward the mean). Bias: mean of predicted minus
true. LoA: half-width of the 95\% limits of agreement, $1.96 \times$ the standard
deviation of the difference. PASTA and Plasma-CycleGAN output intensities in $[0,1]$,
converted to SUVR with a constant fitted on the training set. Per cohort, the best value in each column is in bold and the second best is underlined;
best means closest to 1 for the slope and SD ratio, closest to 0 for the bias and
smallest for the LoA.}\label{tab:calib}
\begin{tabular*}{\textwidth}{@{\extracolsep\fill}lcccc}
\toprule
Model & Slope & SD ratio & Bias & LoA \\
\midrule
\multicolumn{5}{@{}l}{\textit{ADNI} ($n=111$, threshold 1.29)} \\[2pt]
PASTA              & 0.01 & 0.25 & \textbf{$-$0.002} & 0.475 \\
CoCoLIT            & 0.00 & 0.30 & $-$0.218 & 0.485 \\
IL-CLDM            & 0.33 & \underline{1.17} & $-$0.174 & 0.608 \\
Plasma-CycleGAN    & 0.08 & 0.53 & \underline{$-$0.005} & 0.492 \\
\arrayrulecolor{black!35}\cmidrule[0.3pt]{1-5}\arrayrulecolor{black}
SUVR prior only    & \underline{0.43} & \textbf{0.85} & $+$0.055 & \underline{0.433} \\
SUPRA (ours)       & \textbf{0.57} & \textbf{0.85} & $+$0.009 & \textbf{0.357} \\
\midrule
\multicolumn{5}{@{}l}{\textit{A4/LEARN} ($n=143$, threshold 1.32)} \\[2pt]
PASTA              & $-$0.02 & 0.25 & \underline{$-$0.020} & 0.447 \\
CoCoLIT            & 0.05 & 0.30 & $-$0.074 & 0.426 \\
IL-CLDM            & 0.01 & \textbf{0.85} & $+$0.274 & 0.555 \\
Plasma-CycleGAN    & 0.00 & 0.27 & $+$0.045 & 0.442 \\
\arrayrulecolor{black!35}\cmidrule[0.3pt]{1-5}\arrayrulecolor{black}
SUVR prior only    & \textbf{0.33} & \underline{0.73} & \textbf{$+$0.001} & \underline{0.398} \\
SUPRA (ours)       & \underline{0.28} & 0.54 & $-$0.026 & \textbf{0.364} \\
\botrule
\end{tabular*}
\end{table}

\begin{table}[htbp]
\caption{Amyloid status from the predicted cortical SUVR on the test sets: sensitivity and
specificity at the reading threshold (1.29 for ADNI, 1.32 for A4/LEARN), and at the
recalibrated threshold, chosen on the validation set (maximum Youden index). True status: official UC
Berkeley status on ADNI, true cortical SUVR above 1.32 on A4/LEARN. Recalibrated thresholds were computed only for our models. Per cohort, the best value in each column is in bold and the second best is underlined.}\label{tab:status}
\begin{tabular*}{\textwidth}{@{\extracolsep\fill}lccccc}
\toprule
 & \multicolumn{2}{c}{Reading threshold} & \multicolumn{3}{c}{Recalibrated threshold} \\
\cmidrule{2-3}\cmidrule{4-6}
Model & Sensitivity & Specificity & Threshold & Sensitivity & Specificity \\
\midrule
\multicolumn{6}{@{}l}{\textit{ADNI} ($n=111$, threshold 1.29)} \\[2pt]
PASTA              & 0.55 & 0.55 & -- & -- & -- \\
CoCoLIT            & 0.02 & \textbf{0.99} & -- & -- & -- \\
IL-CLDM            & 0.31 & 0.80 & -- & -- & -- \\
Plasma-CycleGAN    & 0.57 & 0.54 & -- & -- & -- \\
\arrayrulecolor{black!35}\cmidrule[0.3pt]{1-6}\arrayrulecolor{black}
SUVR prior only    & \textbf{0.90} & 0.74 & 1.33 & 0.88 & 0.75 \\
SUPRA (ours)       & \underline{0.74} & 0.91 & 1.21 & \underline{0.90} & \underline{0.80} \\
No plasma          & 0.48 & 0.68 & 1.20 & 0.81 & 0.35 \\
\shortstack[l]{No MRI input to the\\refinement network} & 0.67 & \underline{0.93} & 1.22 & \textbf{0.93} & \textbf{0.83} \\
\midrule
\multicolumn{6}{@{}l}{\textit{A4/LEARN} ($n=143$, threshold 1.32)} \\[2pt]
PASTA              & \textbf{1.00} & 0.03 & -- & -- & -- \\
CoCoLIT            & 0.94 & 0.14 & -- & -- & -- \\
IL-CLDM            & \textbf{1.00} & 0.00 & -- & -- & -- \\
Plasma-CycleGAN    & \textbf{1.00} & 0.03 & -- & -- & -- \\
\arrayrulecolor{black!35}\cmidrule[0.3pt]{1-6}\arrayrulecolor{black}
SUVR prior only    & \underline{0.95} & 0.34 & 1.44 & 0.66 & \textbf{0.86} \\
SUPRA (ours)       & \underline{0.95} & \underline{0.49} & 1.41 & \textbf{0.75} & 0.71 \\
No plasma          & 0.87 & 0.37 & 1.42 & 0.57 & 0.57 \\
\shortstack[l]{No MRI input to the\\refinement network} & 0.94 & \textbf{0.51} & 1.40 & \underline{0.72} & \underline{0.80} \\
\botrule
\end{tabular*}
\end{table}

\clearpage

\end{appendices}

\end{document}